# ROI-Aware Multiscale Cross-Attention Vision Transformer for Pest Image Identification

[1]Ga-Eun Kim and [*,1]Chang-Hwan Son

[1]Department of Software Science & Engineering, Kunsan National University

558 Daehak-ro, Gunsan-si 54150, Republic of Korea

[*]Corresponding Author

Phone Number: 82-63-469-8915; Fax Number: 82-63-469-7432

E-MAIL: changhwan76.son@gmail.com; cson@kunsan.ac.kr

**Abstract**

The pests captured with imaging devices may be relatively small in size compared to the entire images, and complex backgrounds have colors and textures similar to those of the pests, which hinders accurate feature extraction and makes pest identification challenging. The key to pest identification is to create a model capable of detecting regions of interest (ROIs) and transforming them into better ones for attention and discriminative learning. To address these problems, we will study how to generate and update the ROIs via multiscale cross-attention fusion as well as how to be highly robust to complex backgrounds and scale problems. Therefore, we propose a novel ROI-aware multiscale cross-attention vision transformer (ROI-ViT). The proposed ROI-ViT is designed using dual branches, called Pest and ROI branches, which take different types of maps as input: Pest images and ROI maps. To render such ROI maps, ROI generators are built using soft segmentation and a class activation map and then integrated into the ROI-ViT backbone. Additionally, in the dual branch, complementary feature fusion and multiscale hierarchies are implemented via a novel multiscale cross-attention fusion. The class token from the Pest branch is exchanged with the patch tokens from the ROI branch, and vice versa. The experimental results show that the proposed ROI-ViT achieves 81.81%, 99.64%, and 84.66% for IP102, D0, and SauTeg pest datasets, respectively, outperforming state-of-the-art (SOTA) models, such as MViT, PVT, DeiT, Swin-ViT, and EfficientNet. More importantly, for the new challenging dataset IP102(CBSS) that contains only pest images with complex backgrounds and small sizes, the proposed model can maintain high recognition accuracy, whereas that of other SOTA models decrease sharply, demonstrating that our model is more robust to complex background and scale problems.

**1. Introduction**

Pests can damage the leaves, stems, and fruit of crops, resulting in reduced productivity and quality. According to the Food and Agriculture Organization, pests cause approximately 40% loss of global crop production annually. Therefore, it is necessary to identify the types of pests that occur in the field and perform prompt and timely control. Traditionally, pests have been identified by experienced farmers and experts; however, this is quite costly and time-consuming because pests are diverse and have similar appearances in shape and texture, with low discrimination for some species. Additionally, exotic pests occur because of the increase in international trade, making pest identification more difficult. To this end, stationary and handheld crop-monitoring systems with built-in cameras, such as digital traps[1] and mobile pest-monitoring systems[2], have recently been developed for automatic pest identification. Such crop-monitoring systems collect massive pest image datasets and train machine learning algorithms (ML) for decision-making and prediction.

Before the advent of deep learning (DL), handcrafted feature descriptors such as SIFT[3], HOG[4], and color histograms [5] were mainly used, and a support vector machine (SVM)[6] was applied for classifier learning. However, these handcrafted feature extractors have limitations in generating abundant and deep features and are only capable of shallow learning; therefore, they have not reached satisfactory performance levels. Nowadays, ML software employs deep learning that automatically produces discriminative and deep features from raw data. Early DL models followed Convolutional Neural Network (CNN)-based frameworks such as ResNet[7], EfficientNet[8], DenseNet[9], and FPN[10]; however, vision transformers (ViTs) with self-attention and positional encoding have become popular. Their performance has improved rapidly and is now comparable to that of the CNN family. The representative ViT models include DeiT[11], Swin-ViT[12], PVT[13], and MViT[14]. Following these trends, variants of CNN and ViT[15,16,17,18,19] and lightweight versions[20,21] optimized for pest identification, detection, and counting have been developed in agricultural fields.

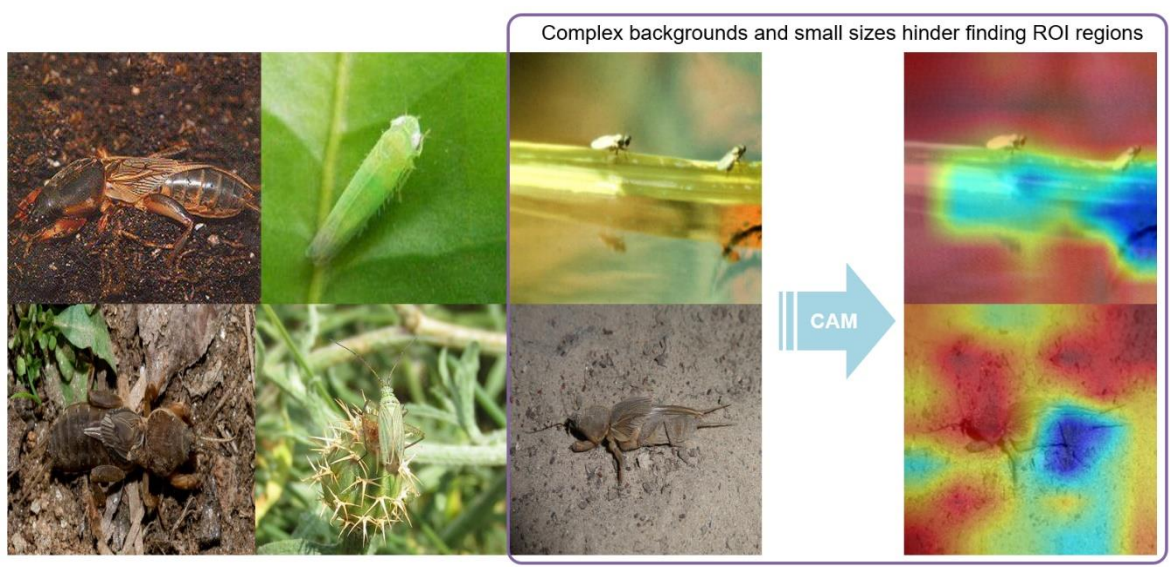

Fig. 1. Main issues of pest image identification: complex backgrounds and small sizes.

**1.1 Main issue for pest image identification**

Despite recent progress in the CNN and ViT families, pest image identification still suffers from problems such as complex backgrounds [22,23,24,25,26] and small sizes [2,17,27]. For accurate pest image identification, it is crucial to separate regions of interest (ROIs), such as wings and heads, from the background, which contains clues for pest identification. In particular, complex backgrounds with textures and colors similar to those of pests interfere with ROI detection, making the features indistinct and weakening self-attention. For pests that are small compared with the entire image, it is difficult to pool ROI features and strengthen sparsity. Moreover, inter- and intra-class variations in pests prevent discriminative feature learning.

Fig. 1 shows examples of why pest image identification is challenging because of cluttered backgrounds and small sizes. In the rightmost part of Fig. 1, an example of saliency maps is provided, where red colors indicate the most noticeable areas for pest identification, whereas areas with blue colors are relatively unimportant. In these maps, pests are not recognized as salient areas; however, the backgrounds are more focused. These saliency maps support our assumption that complex backgrounds and small scales interfere with finding ROIs, ultimately reducing pest identification accuracy. These problems are not only limited to pest image identification but also appear in crop disease recognition.

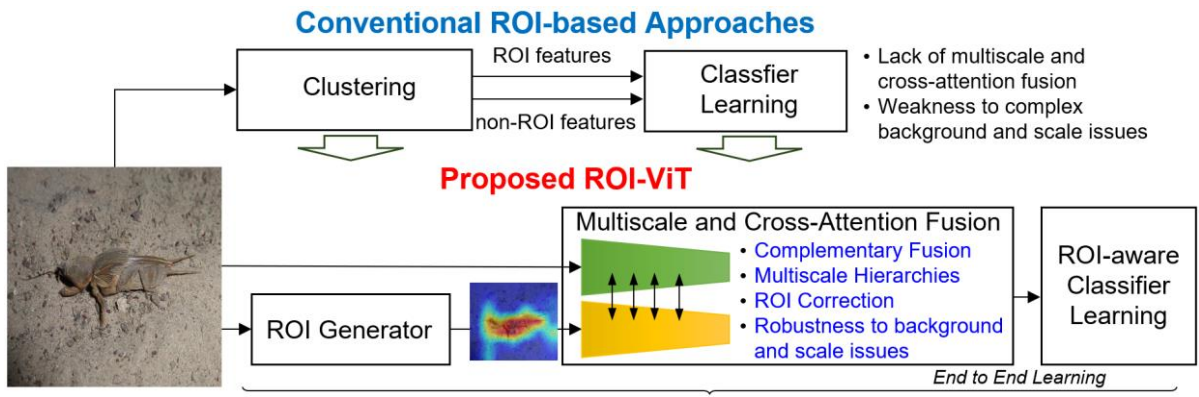

Fig. 2. Proposed ROI-ViT model with complementary feature fusion, multiscale hierarchies and ROI correction functions to overcome complex background and scale problems.

**1.2 ROI-based approaches in agriculture**

Over the past few decades, ROI-based approaches have evolved in agriculture to address these issues. Pest information exists in the ROI and not in the background. Therefore, it is important to identify the ROIs in pest images. ROIs can be expressed in various ways; however, clustering[23,24], superpixels[25], and saliency[22] have been commonly used. In the initial stage, clustering algorithms, such as k-means clustering, were applied to input pest images for ROI and background separation[23,23], which makes the grouped features more discriminative. This approach is simple, yet very powerful for increasing accuracy. This concept has been adopted in DL frameworks. For grape-disease identification, superpixels, defined as the perceptual grouping of pixels with similar visual properties, are concatenated with input crop images to provide ROI information[25]. However, this is still an early approach. In other words, intermediate and later fusions cannot be implemented. In addition, ROI detection and CNN learning were performed separately. Recently, a more advanced LSA-Net [28] was introduced for leaf disease identification to integrate ROI prediction into the VGG backbone in an end-to-end manner. An image segmentation branch was added to predict the ROIs, and both early and late fusions were achieved via simple concatenation, thereby resulting in a significant improvement in the recognition accuracy. This approach is regarded as a one of CNN-based attention model. Further details on the ROI-based models are discussed in related works section.

**1.3 Proposed approach**

Inspired by a previous study[28], this study presents a method for building a new ViT that leverages ROI maps and transforms them into better ones for complementary feature fusion. In particular, for a new challenging dataset that contains only pest images with complex backgrounds and small sizes, we demonstrate that the proposed ROI-ViT is superior to conventional state-of-the-art (SOTA) models. Fig. 2 shows the main difference between conventional ROI-based approaches and the proposed ROI-ViT. As mentioned in the previous section, conventional ROI-based approaches divide features into ROI and non-ROIs. This is an effective way to enhance the discriminating power of the features, but it has drawbacks. First, the unsupervised clustering is inaccurate. Second, the two steps of clustering and classifier learning are separated, making learning inconvenient. More advanced ROI-based CNNs [22,25,28] significantly improve accuracy; however, no new design has been presented for the implementation of complementary multiscale-feature fusion at intermediate layers. This means that powerful features, such as multiscale hierarchies, ROI correction, and cross-attention fusion, are absent. In addition, conventional CNNs [15,16,18,22,25,28] and ViTs [10-14] have not demonstrated their capability to overcome both complex background and scale problems.

To solve these issues, in the proposed ROI-ViT, a new ROI generator is constructed using soft segmentation and a class activation map (CAM) [29] for more accurate ROI detection. In addition, a novel ROI-ViT backbone is designed using dual branches with multiscale and cross-attention fusion to achieve complementary feature fusion, multiscale hierarchies, and ROI correction. The proposed multiscale cross-attention fusion has the powerful capability to transform incorrect ROI maps into better ones through multiple stages and focus on more salient regions, thereby enabling our model to be robust to complex background and scale problems. This is the strength of our model, which differs from conventional ROI-based CNNs and ViTs. In Fig. 2, the ROI-aware classifier is an affine layer that maps complementary multiscale features into predictions. All steps of the ROI generator, multiscale cross-attention fusion, and classifier learning are integrated and then trained in an end-to-end manner.

**1.4 Contributions**

- In this study, a novel ROI-ViT modified for pest image identification is introduced. In particular, there are still challenging problems, such as complex backgrounds and small sizes, that hinder pest identification. To address these problems, two types of ROI generators using soft segmentation and CAM are introduced. Dual branches are designed with multiscale and cross-attention fusion to achieve complementary fusion, multiscale hierarchies, and ROI correction. Throughout the experiments, we verified that the proposed ROI-ViT can focus on more salient areas via ROI updating and achieve accurate pest identification even in hard samples with complex backgrounds and small sizes. To the best of our knowledge, this is the first attempt to integrate ROIs into the ViT for complementary multiscale-feature fusion. FNSTC [18] has dual branches similar to our model, but with different objectives. The FNSTC aims to solve the class imbalance problem, whereas our goal is to overcome complex background and scale problems. In addition, in FNSTC, the ROI generator and cross-attention fusion are excluded. The CNN and ViT backbones are combined to leverage their local and global feature learning properties.

- For public pest image datasets, such as IP102, D0, and SauTeg, the proposed ROI-ViT can achieve 81.81%, 99.64%, and 84.66% recognition accuracy, respectively, which is the best among SOTA models, such as Swin-ViT, MViT, PVT, and EfficientNet. More importantly, for a new dataset that only contains pest images with complex backgrounds and small sizes, the proposed ROI-ViT still maintains high accuracy, whereas the accuracies of existing SOTA models drop sharply. This result indicates that the proposed ROI-ViT is much more robust to complex backgrounds and small sizes than conventional SOTA models. Therefore, our model has the potential to be used as a baseline for pest image identification. Evaluation scores can also be used to compare performance. *Our codes and dataset is available from https://github.com/cvmllab.*

- The proposed ROI-ViT is an extremely advanced version of LSA-Net [28]. Therefore, major improvements have been made in this regard. In LSA-Net, simple concatenation is used for attention, and only early and later fusions exist. However, this may not be sufficient for complementary feature fusion. In addition, a multiscale approach is not implemented, indicating that this model is too weak to scale problems. The backbone used is outdated. Most importantly, the LSA-Net targets leaf disease identification. In other words, it has not yet been optimized for pest identification. However, in the proposed ROI-ViT, multiscale hierarchies and cross-attention are incorporated to fuse multiscale ROI and non-ROI features at the intermediate stages, thereby progressively updating the ROIs. This indicates that the proposed multiscale cross-attention fusion is more sophisticated than the concatenation-based fusion. In addition, modern ViTs are adopted to build a dual branch and two types of ROI generators are tested.

**2. Related Works**

Challenging problems in pest image identification can be broadly divided into three categories: small size, long-tailed distribution (LTD) and complex backgrounds. As shown in Fig. 1, complex backgrounds and small sizes can make feature extraction inaccurate. The LTD indicates that the pest image datasets may have long-tailed distributions [18]. For example, the large-scale IP102 dataset contains a large number of examples for most species; however, there are fewer examples for some species. LTD is considered a data imbalance problem that has a negative impact on the loss calculation. Therefore, in this section, we introduce various approaches for addressing these problems.

**2.1 Multiscale design**

Scaling is a long-standing problem in image classification and object detection. Even the same object may have different sizes, depending on the shooting distance and viewing angle. Multiscale image representations, such as the wavelet transform[30] and Laplacian pyramid[31], have been steadily adopted to address this problem. In particular, attempts have been made to incorporate these multiscale representations into DL frameworks. Representative models include the FPN[10], BiFPN[8], and UNet[32]. Another research direction involves utilizing or modifying standard convolution operations. A simple approach involves applying a variety of filters of different sizes[33] and constructing multiple branches[34] for a wide range of receptive fields. Deformable[35] and dilated convolutions[36] have been devised and have been successful in geometric deformation modelling and contextual understanding. Other approaches are based on super-resolution[37] and bounding box selection[38].

These multiscale approaches have been applied directly and slightly modified to make them more suitable for pest image identification. Multiscale-feature extraction is designed using dilated convolution[15] to generate multiple scales. Multiple branches[16] with different kernel sizes are constructed using DenseNet to adjust the receptive field sizes. These are simple models that apply only convolution filters. In addition, ClusRPN[17], a variant of RPN, is proposed to output the location candidates of cluster regions and apply local detectors separately. Although this model is effective for extremely small-scale datasets such as aerial images and wheatears, its implementation appears to be complicated and its performance is sensitive to the object density level. For a handheld pest-monitoring system[2], a rice planthopper search network (RPSN) with a sensitive score matrix is proposed. However, its architecture is borrowed from the Faster R-CNN [39], and there is only a slight difference between them. Note that RPN families such as RPSN and ClusRPN aim at object detection to localize bounding boxes, which is different from our goal of pest image identification.

**2.2 LTD solutions**

In field environments, LTD phenomena are likely to occur in most pest datasets, because the distribution of pest classes is significantly uneven and depends on different regions, periods, and meteorological conditions. LTD poses a challenge in avoiding biases toward head classes and determining an optimal boundary decision for tail classes[18]. A simple strategy is to apply a data augmentation technique for data balancing. There are more elegant approaches for resolving this issue. Hard example mining is suggested by searching for hard samples and using them in network training. This hard example of mining is adopted in RCNN[39]. Focal Loss[40] and its variants, for example I-confidence loss[38], have been devised to reduce the weights of easily classified samples and focus on harder samples. Recently, meta-learning[41] and decoupled learning[42] have become popular for LTD. Although these models can improve the model performance on tail classes, it is difficult to design algorithms and training models. Therefore, for pest image datasets with LTD, data augmentation and loss modification remain mainstream because of their simplicity and efficiency. More recently, ViT and CNN have been combined to leverage global and local feature learning[18]. That is, ViT performs well for head classes, and CNN performs much better for tail classes.

**2.3 ROI-aware CNNs and ViTs**

Attention models have become mainstream for image classification owing to their effectiveness in extracting semantic information and reducing the influence of complex backgrounds. Early models focused on designing spatial and channel attention maps within CNN frameworks. Commonly used models include SENet [43], CBAM[44], and SAN[45]. Another research direction involves the development of ROI-aware CNNs. Here, ROIs can be interpreted differently; however, in most cases, superpixels[25], saliency[22], and segmentation[23,24,28] have been adopted to describe the ROIs. To overcome complex backgrounds, a superpixel-based CNN[25] is proposed by stacking superpixels and crop image for graph disease identification. This model is considered to be an early fusion model. A saliency map[22] is also generated based on the global-contrast and graph-cut algorithm for ROI-aware pest identification. However, saliency prediction and CNN training remain separate; thus, they are inefficient. To improve this, LSA-Net[28] is proposed to integrate ROI prediction into the VGG backbone, where early and later fusions are applied and end-to-end learning is achieved. This architectural design significantly increased the leaf identification accuracy.

Recently, attention has shifted from ROI-aware CNNs to ViTs. Originally, Transformer[46] was developed for the natural language processing (NLP) field. It is a convolution-free model that relies only on a self-attention mechanism to draw global dependencies between tokens. Inspired by the success of Transformer, ViTs, a variant of Transformer used for vision tasks, have been actively studied. The first ViT[47] is the full-transformer model directly applied for image classification, which splits the input image into tokens (i.e., image patches) of a fixed size and applies transformer layers for self-attention. However, it is challenging to directly apply ViT to dense prediction tasks because its output map is single-scale and has a low resolution. Multiscale ViTs have been investigated to overcome these drawbacks. Popular models include PVT[13] and MViT[14], which progressively expand channel resolution through several stages, while reducing spatial resolution. This strategy allows for the creation of a multiscale pyramid of features. Local attention[48] and cross attention [49] are actively being studied to achieve local group attention and complementary feature fusion. The aforementioned ViT models are being rapidly adopted for pest image identification. However, in most cases, ViTs are used directly [50] or slightly modified for multilabel classification[51], hybrid architecture [18], and multimodal fusion [52].

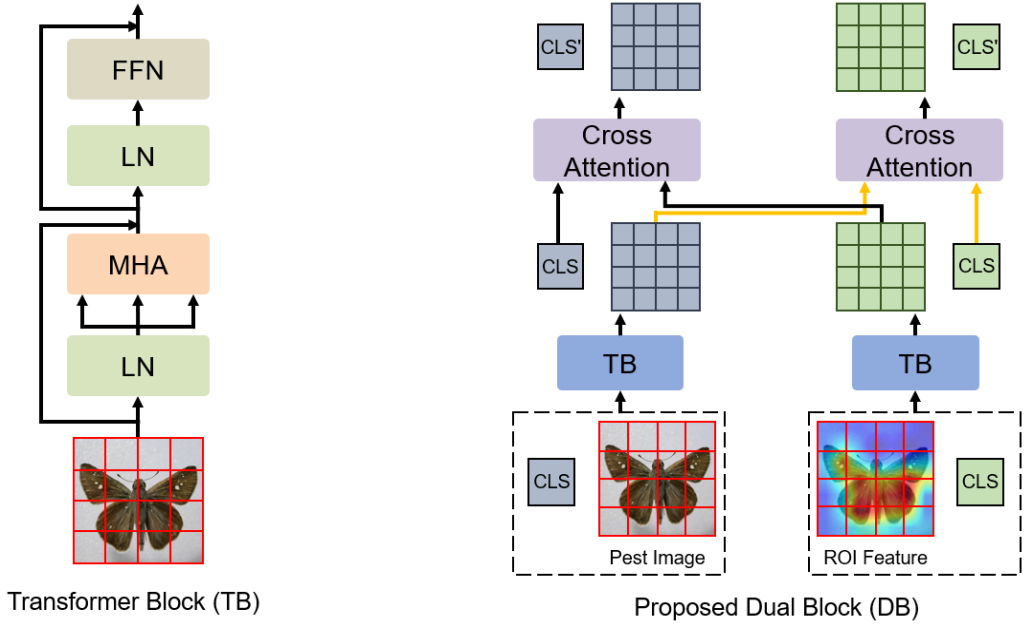

Fig. 3. Conventional transformer block (left) vs. proposed dual block based on cross-attention fusion (right).

## 3. Overview of ViT

The core component of ViTs is the transformer encoder, composed of a sequence of blocks, as shown in the left image of Fig. 3, where each transformer block (TB) contains a multihead attention (MHA) and feed-forward network (FFN). In particular, MHA models long-range dependencies among input tokens and exhibits great flexibility. Layer normalization (LN) is applied before each block, and a residual skip connection (RSC) after every block. Technically, given the input feature $x(0)$, where the value in the brace is the block index, the $k$-th TB operates as follows:

$$x(0) = [x_{cls}||x_{patch}] + x_{pos} \qquad (1)$$

$$y(k) = x(k-1) + MHA\left(LN(x(k-1))\right) \qquad (2)$$

$$x(k) = y(k) + FFN\left(LN(y(k))\right) \qquad (3)$$

where $x_{patch} \in \mathbb{R}^{N \times C}$ is the patch token and $x_{cls} \in \mathbb{R}^{1 \times C}$ is a global token, also called classification token (CLS), which interacts with all patch tokens at every TB to form a global feature representation. $N$ and $C$ are the number of patch tokens and embedding dimensions, respectively. The symbol || represents the concatenation operation. According to Eq. (1), ViT first divides an input image into fixed-size patch tokens ($x_{patch}$) and stacks the CLS token on top of these tokens. Position embedding ($x_{pos}$) is then added to compensate for missing positional information during tokenization. Subsequently, the input feature $x(0)$ is sequentially passed through the MHA and FFN, executing the TB once, as expressed in Eqs. (2) and (3). This process is repeated multiple times, and finally, the CLS token is used for classification.

## 4. Proposed ROI-ViT

As pest information exists only in the ROI and not in the background, ROIs play an important role in pest identification. However, complex backgrounds and small objects are the main obstacles in ROI separation and discriminative feature learning. To the best of our knowledge, advanced ViTs that can overcome these problems in pest image identification have not yet been fully explored. However, in-depth analyses and evaluation comparisons are lacking. Motivated by this, our proposed approach seeks to leverage the advantages of ROIs and presents a novel architectural design for integrating the ROI generator, multiscale hierarchies, and cross-attention fusion into the ROI-ViT.

### 4.1. Proposed dual block vs. conventional TB

Unlike TB, the proposed ROI-ViT has dual blocks (DB), as shown in the right part of Fig. 3, where two different types of images, that is, pest image and ROI map, are taken as input and fed into a series of blocks that arrange TB and the proposed cross-attention fusion module. In particular, in this module, two types of CLS tokens are updated by fusing patch tokens from other branches, thereby enabling complementary feature fusion. This leads to an improved global feature representation, which is helpful in increasing the accuracy of pest identification. In summary, DBs are responsible for cross attention to update CLS tokens from dual branches, whereas TBs are responsible for self-attention to update the patch and CLS tokens from each branch.

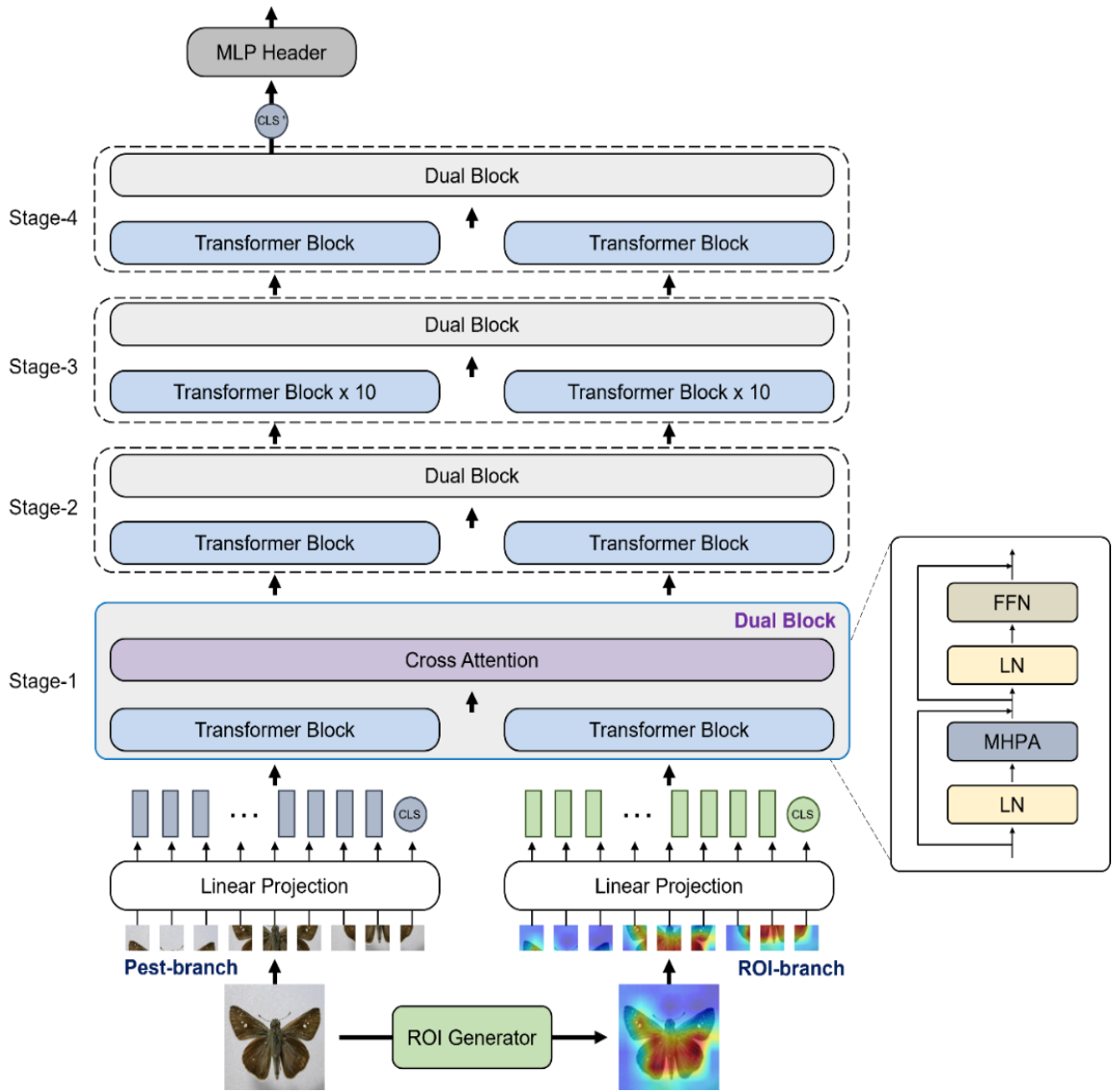

Fig. 4. Architecture of the proposed ROI-ViT for pest image identification.

**4.2. Proposed architecture**

Fig. 4 illustrates the architecture of the proposed ROI-ViT. Our model mainly consists of an ROI generator, ROI-ViT backbone, and MLP Header. Notably, the ROI-ViT backbone is used as a dual feature extractor for complementary multiscale-feature fusion. The proposed backbone has two characteristics: cross-attention fusion and multiscale-feature hierarchies.

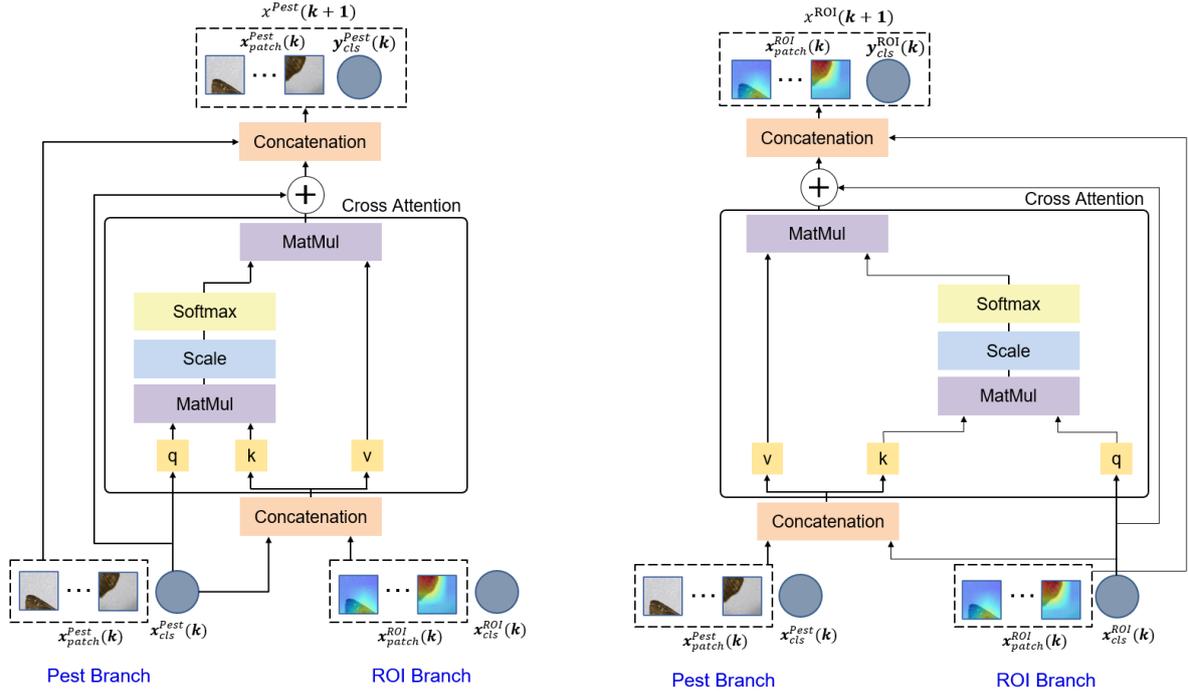

Fig. 5. Proposed cross-attention fusion module for Pest branch (left) and ROI branch (right).

*4.2.1. Cross-attention Fusion.*

To fuse the complementary multiscale features, a dual branch is constructed, as shown in Fig. 4. One is Pest branch and the other is ROI branch. In the Pest branch, pest multiscale features are extracted from the input pest patch tokens, and in the ROI branch, the input ROI patch tokens are transformed into ROI multiscale features. In the dual branch, two types of multiscale features are generated. More importantly, each branch passes through the proposed cross-attention fusion module for complementary multiscale-feature fusion. In the cross-attention fusion module, two types of multiscale features are used to update the CLS tokens, thereby enhancing the global feature representation. For reference, the CLS token is fed into the MLP head for classification. No other patch tokens are used in this study.

The proposed cross-attention fusion module for the Pest and ROI branches is illustrated in Fig. 5. To facilitate the understanding of the proposed cross-attention fusion, we describe the procedure only for the Pest branch. The same procedure applies to the ROI branch, except that the CLS and patch tokens are only swapped from another branch. The final goal of cross attention is to update the CLS token of the Pest branch and leave the other patch tokens unchanged. To achieve this, the CLS token of the Pest branch serves as a query to interact with the patch tokens from the ROI branch through cross-attention. Specifically, for the Pest branch, the patch tokens from the ROI branch are collected and concatenated with the CLS token of the Pest branch.

$$x'^{ROI}(k) = [x_{cls}^{Pest} || x_{patch}^{ROI}] \qquad (4)$$

where $x^i$ indicates the token sequence at branch $i$, and $i$ is $Pest$ or $ROI$ for the Pest and ROI branches, respectively. $x_{cls}^i$ and $x_{patch}^i$ represent the CLS and patch tokens of branch $i$, respectively. In Eq. (4), $k$ denotes the block index for the TB and DB. For brevity, the block index $k$ is omitted but may be specified depending on the context. Eq. (4) means that the CLS token ($x_{cls}^{pest}$) of the Pest branch is concatenated with the patch tokens ($x_{patch}^{ROI}$) of the ROI branch to form a new patch token $x'^{ROI}$. Given the $x'^{ROI}$, the proposed cross-attention is calculated as follows,

$$q = x_{cls}^{Pest} W_q, \qquad k = x'^{ROI} W_k, \qquad v = x'^{ROI} W_v \qquad (5)$$

$$A = softmax\left(\frac{qk^T}{\sqrt{C/h}}\right), \qquad CA(x'^{ROI}) = Av \qquad (6)$$

where $W_q$, $W_k$, and $W_v \in \mathbb{R}^{C \times C}$ are learnable parameters, and $C$ is the embedding dimension. In Eq. (5), the CLS token of the Pest branch $x_{cls}^{Pest}$ is used as a query to calculate the similarity matrix $A \in \mathbb{R}^{1 \times N}$. To normalize the matrix elements, $softmax$ is applied.

$$y_{cls}^{Pest} = x_{cls}^{Pest} + CA(x'^{ROI}) \qquad (7)$$

$$x^{Pest}(k+1) = [y_{cls}^{Pest} || x_{patch}^{Pest}] \qquad (8)$$

The updated patch token $CA(x'^{ROI})$ via cross-attention is added to the input CLS token $x_{cls}^{Pest}$, thereby producing a new CLS token $y_{cls}^{Pest}$. This token is combined with its corresponding patch token $x_{patch}^{Pest}$ to form a new pest-token sequence $x^{Pest}(k+1)$. Again, only the CLS token is updated, as shown in Eq. (8) to enhance the global feature representation. For simplicity, $LN$ is omitted from Eq. (7). In this study, $LN$ is conducted before applying the $CA$ but $FFN$ is not applied after $CA$, which differs from TB.

*4.2.2 Multiscale-Feature Hierarchies*

Multi-head pooling attention (MHPA) [14] is adopted for multiscale-feature hierarchies. MHPA is a variant of the MHA that allows for flexible resolution. In contrast to the original MHA, where the feature resolution remains fixed, MHPA can reduce the space-channel dimension by applying a pooling operator. It should be noted that in the proposed backbone shown in Fig. 4, MHA is replaced by the MHPA to realize multiscale-feature hierarchies. This implies that as the layer depth increases, the channel capacity expands progressively while reducing the spatial resolution. The early layers learn low-level features such as edges and textures, whereas the deeper layers focus on deriving high-level visual information. This type of multiscale approach has proven beneficial for transformer models across a variety of visual recognition tasks.

Following MHA, MHPA [14] first projects the input token sequence $x^i(k)$ into an embedding space via a matrix product as follows:

$$q = \mathbb{P}_q(x^i W_q), \qquad k = \mathbb{P}_k(x^i W_k), \qquad v = \mathbb{P}_v(x^i W_v) \qquad (9)$$

where $x^i(k)$ denotes the token sequence of $k$-th block for the Pest and ROI branches. For brevity, block index $k$ is omitted; however, it is specified when necessary. In Eq. (9), the MHPA applies a pooling operator ($\mathbb{P}$) for scale conversion, which differentiates it from MHA. The pooling operator is implemented using a convolution operator that scales using the stride factor.

Next, a similarity matrix to determine the correlation between patch tokens is obtained using query($q$), key($k$), and value($v$), as follows:

$$Attention(\boldsymbol{q}, \boldsymbol{k}, \boldsymbol{v}) = softmax\left(\frac{\boldsymbol{q}\boldsymbol{k}^T}{\sqrt{C/h}}\right)\boldsymbol{v} \tag{10}$$

$$\boldsymbol{x}^i(k+1) = MHPA(\boldsymbol{x}^i(k)) = Attention(\boldsymbol{q}, \boldsymbol{k}, \boldsymbol{v}) + \mathbb{p}_q(\boldsymbol{x}^i(k)) \tag{11}$$

where $\boldsymbol{qk}^T$ is the matrix used to measure the similarity between patch tokens. $h$ is the number of heads, and the other symbols are the same as in the cross-attention fusion. Regardless of the branches, the MHPA applies to TBs in the same manner as mentioned above. Once again, multiscale-feature hierarchies are achieved through MHPA, which means that TB is responsible for scale conversion.

Table 1 summarizes the scale changes according to the stage. As shown in Fig. 4, a stage is defined as a DB or a set of TBs and DB that operate on the same scale with identical resolution across the channel and spatial dimensions. From Table 1, it can be seen that as the stage increases, the spatial resolution decreases, but the channel capacity expands. Specifically, if we downsample the spatial resolution by 4x, we increase the channel dimension by 2x. This approximately preserves the computational complexity across stages [14].

Table 1. Scale conversion, according to stages.

| **Stages** | **Output patch token sizes** | **Output CLS token sizes** |
|---|---|---|
| Linear Projection | $\frac{H}{4} \times \frac{W}{4} \times C$ | $1 \times C$ |
| Stage-1 | $\frac{H}{4} \times \frac{W}{4} \times C$ | $1 \times C$ |
| Stage-2 | $\frac{H}{8} \times \frac{W}{8} \times 2C$ | $1 \times 2C$ |
| Stage-3 | $\frac{H}{16} \times \frac{W}{16} \times 4C$ | $1 \times 4C$ |
| Stage-4 | $\frac{H}{32} \times \frac{W}{32} \times 8C$ | $1 \times 8C$ |

*4.2.3 ROI Generator*

The ROI generator initializes the ROI map from the input pest image to be fed into the proposed ROI-ViT backbone. This initial ROI helps the backbone to locate spatially more important areas for pest identification.

The ROI map is considered a spatial attention map. In this study, the ROI maps are defined as soft segmentation and CAM. We will conduct an experiment later to determine which ROIs are better suited for pest identification.

First, CAM[29] is a technique for identifying discriminative regions using a linearly weighted combination of the activation maps of the last convolutional layer. CAM creates a localized visual explanation; however, its architecture requires global averaging pooling and is sensitive to gradient information. Therefore, this study adopts Score-CAM[29], a gradient-free approach that has been shown to achieve better visual performance and fairness for interpreting the decision-making process. First, Score-CAM generates an ROI map, that is, a saliency map, using the following equation:

$$ROI = L_{Score-CAM}^{c} = ReLU\left(\sum_{k} a_{k}^{c} A_{l}^{k}\right) \tag{12}$$

where $A_l^k$ denotes the $k$-th activation map of layer $l$ and $a_k^c$ is the corresponding weight, and $c$ denotes the target class node in the prediction layer. $ReLU$ is a rectified linear unit function used to consider only features that have a positive influence on the target class. Therefore, $L_{Score-CAM}^{c}$ is viewed as a heatmap representing the extent to which each activation map affects the target class $c$. The weight $a_k^c$ is calculated as follows:

$$H_l^k = NF\left(UP(A_l^k)\right) \tag{13}$$

$$s_k^c = f^c(X^{pest} \circ H_l^k) \tag{14}$$

$UP$ is an operation that upsamples $A_l^k$ into the input image size. $NF$ is a normalization function that maps the input matrix to [0,1]. In Eq. (14), the symbol ∘ represents Hadamard product and $f^c$ is the prediction score of target class c given the model $f$ and input image $X^{pest}$. This equation indicates that $H_l^k$ is used as a mask to measure the extent to which the masked input $X \circ H_l^k$ affects target class prediction. Finally, $s_k^c$ is converted into the weight value $a_k^c$ of [0–1] via the SoftMax function.

$$a_k^c = softmax(s_k^c) \tag{15}$$

Second, soft segmentation produces predicted maps to separate the input pest image into the background and ROI. The predicted map resembles a B/W image and has a value of [0–1]. In LSA-Net [28], the integration of the image segmentation module into the VGG backbone led to a significant improvement in leaf disease identification. Based on this, an image segmentation module is considered for the ROI generator.

$$ROI = f_{seg}(X^{pest}) \tag{16}$$

where $f_{seg}$ denotes the image segmentation model. There are differences between LSA-Net [28] and the proposed ROI-ViT. In LSA-Net, only the early and later fusions are implemented. In addition, multiscale-feature hierarchies have not been considered. However, in the proposed ROI-ViT, multiscale cross-attention fusion is incorporated for complementary multiscale-feature fusion, thereby producing more discriminative and enhanced multiscale features.

*4.2.4 MLP Header*

In Fig. 4, the MLP Header is an ROI-aware classifier that leverages fused ROI features, in contrast to conventional ViTs[11-14]. The proposed ROI-ViT backbone generates two types of tokens via dual branches. However, only CLS tokens, excluding patch tokens, are used for classification. Through multiple stages, the CLS tokens summarize all the patch tokens as global feature representations. It is also empirically demonstrated through experiments that using the CLS token of the Pest branch yields better accuracy than using two CLS tokens of a dual branch. In other words, only the CLS token of the Pest branch is fed into the MLP Header. In this study, a fully connected layer is used as the MLP Header.

Table 2. Public pest image datasets.

| Datasets | Classes | Training | Test | Total |
|---|---|---|---|---|
| IP102 [53] | 86 | 24,727 | 10,569 | 35,296 |
| D0 [54] | 40 | 3,139 | 1,368 | 4,507 |
| SauTeg [55] | 43 | 3,310 | 1,444 | 4,754 |

## 5. Experimental Results

### 5.1. Datasets and experimental setting

In this study, three public pest image datasets, IP102[53], D0[54], and SauTeg[55], were used. Table 2 summarizes the dataset information, including the total number of classes and the number of training and test datasets. The IP102 dataset contained 102 classes. However, this dataset contained larvae that were not target images. In addition, there were only a small number of images for some species. However, there are LTD problem that is not our goal. Moreover, some images contained artificially generated watermarks. Therefore, in this study, we filtered out inappropriate images and selected 86 pest classes. The D0 and SauTeg datasets contain 4,507 and 4,754 pest images belonging to 40 and 43 classes, respectively. All datasets were subdivided into training and test datasets at a 7:3 ratio. **It should be noted that, in addition to these datasets, we will construct a new test dataset called IP102(CBSS) that only contains pest images with complex backgrounds and small sizes. This new dataset verifies that the proposed ROI-ViT is much more robust to complex background and scale problems than conventional SOTA models.**

All images were resized to 224 × 224 pixels for training and testing. The batch size was 10 and the number of epochs was 80. Adam was used as the optimizer and the learning rate was set to 0.0001. PyTorch was used as the deep learning framework. For all the comparison models, the training settings, such as epochs and batches, were the same, except for the learning rate. In this study, the CNN- and ViT-based SOTA models were compared. CNN models include MobileNet(v3)[56], ResNet50[7], ConvNeXt[57] and EfficientNet[8], while ViT models include the Swin-ViT[12], DeiT[11], Cross-ViT[49], PVT[13], MPViT[58] and MViT[14]. All experiments were conducted on an AMD Ryzen 5 5600X 6-Core Processor @ 3.7 GHz with an Nvidia GeForce RTX 3080.

*Our source codes and datasets will be available from https://github.com/cvmllab.*

Fig. 6. Illustration to calculate TP, TN, FN, and FP for a specific C1 class from a given confusion matrix.

**5.2. Quantitative evaluation**

To measure the performance of pest image identification, commonly used precision, recall, F1-score and accuracy were tested.

$$Accuracy = \frac{TP + TN}{TP + TN + FP + FN} \qquad (17)$$

$$Precision = \frac{TP}{TP + FP} \qquad (18)$$

$$Recall = \frac{TP}{TP + FN} \qquad (19)$$

$$F1 = 2 * \frac{Precison * Recall}{Precision + Recall} \qquad (20)$$

where TP is the number of true positives (samples that are correctly identified as pests) and FN is the number of false negatives (samples that are incorrectly identified as the background). TN is the number of true negatives, that is, the samples that are correctly identified as the background, and FP is the number of false positives, that is, the samples that are incorrectly identified as pests. To help you understand, an example is provided in Fig. 6. Given a confusion matrix representing the prediction summary in matrix form, for a specific C1 class, the precision and recall scores were calculated from the first column and row, respectively. Specifically, the precision is the ratio of the TP to the sum of the first column, and the recall is the ratio of the TP to the sum of the first row. The accuracy is the ratio of TP and TN to the total number of test samples. For reference, TN is the remaining part of the confusion matrix, excluding the row and column of the C1 class. For the other classes, the same procedure was conducted to obtain evaluation scores, which were then averaged for all classes.

Tables 3–5 show the accuracy, recall, precision and F1-scores of the IP102, D0, and SauTeg datasets, respectively. As shown in Table 3, the performances of the ViT family, including PVT, Cross-ViT, and MPViT, are comparable to that of EfficientNet, a member of the CNN family. It was also observed that MViT outperformed the conventional CNN and ViT models for the IP102 and SauTeg datasets. For the D0 dataset, MViT ranked second after PVT, indicating that multiscale-feature hierarchies are effective for improving pest image identification. However, MViT scored lower on all metrics than the proposed ROI-ViT. Although MViT is rated as the best among the SOTA models, it may still suffer from finding the ROIs. However, the proposed method leverages the initial ROI information and updates it through cross-attention fusion. In addition, in the TB block, multiscale-feature hierarchies were achieved using MHPA, and complementary feature fusion was performed in the DB block. This results in a greater focus on the ROI, making pest identification more accurate.

For the D0 dataset, the resulting accuracy of the FNSTC is taken from [18], as shown in Table 4. Although FNSTC has dual branches similar to our ROI-ViT for complementary feature fusion, there are major differences between them. First, the goal of the FNSTC is to solve the LTD problem, whereas ours is to overcome complex background and scale problems. Second, the FNSTC combines two types of existing CNN and ViT backbones to learn the local and global features. However, the proposed ROI-ViT presents a new architecture that enables ROI generation, multiscale hierarchy, and complementary multiscale-feature fusion. Owing to these powerful functions, the proposed ROI-ViT can obtain better results than FNSTC.

Table 3. Quantitative evaluation for IP102 datasets.

| Models | Accuracy | Recall | Precision | F1-Score |
|---|---|---|---|---|
| MobileNet(v3) [56] | 71.00% | 0.583 | 0.642 | 0.611 |
| ResNet50 [7] | 71.30% | 0.592 | 0.625 | 0.608 |
| ConvNeXt [57] | 72.14% | 0.499 | 0.589 | 0.540 |
| EfficientNet [8] | 79.01% | 0.656 | 0.713 | 0.683 |
| Swin-ViT [12] | 73.32% | 0.490 | 0.588 | 0.535 |
| DeiT [11] | 77.59% | 0.594 | 0.682 | 0.635 |
| CrossViT [49] | 76.10% | 0.537 | 0.598 | 0.566 |
| PVT [13] | 76.24% | 0.566 | 0.638 | 0.600 |
| MPViT [58] | 76.62% | 0.546 | 0.673 | 0.603 |
| MViT [14] | 80.34% | 0.639 | 0.724 | 0.679 |
| **Proposed ROI-ViT (Segmentation)** | **81.61%** | **0.694** | **0.765** | **0.728** |
| **Proposed ROI-ViT (CAM)** | **81.81%** | **0.700** | **0.743** | **0.721** |

Table 4. Quantitative evaluation for D0 datasets.

| Models | Accuracy | Recall | Precision | F1-Score |
|---|---|---|---|---|
| MobileNet(v3) [56] | 96.63% | 0.960 | 0.965 | 0.963 |
| ResNet50 [7] | 96.92% | 0.959 | 0.963 | 0.961 |
| ConvNeXt [57] | 98.10% | 0.885 | 0.890 | 0.888 |
| EfficientNet [8] | 98.39% | 0.975 | 0.975 | 0.975 |
| Swin-ViT [12] | 97.36% | 0.954 | 0.968 | 0.961 |
| DeiT [11] | 98.24% | 0.977 | 0.979 | 0.978 |
| CrossViT [49] | 98.09% | 0.972 | 0.979 | 0.975 |
| PVT [13] | 99.31% | 0.990 | 0.993 | 0.991 |
| MPViT [58] | 98.83% | 0.983 | 0.985 | 0.984 |
| MViT [14] | 99.27% | 0.991 | 0.991 | 0.991 |
| FNSTC[18] | 98.50% | - | - | - |
| **Proposed ROI-ViT (Segmentation)** | **99.42%** | **0.991** | **0.993** | **0.992** |
| **Proposed ROI-ViT (CAM)** | **99.64%** | **0.992** | **0.994** | **0.993** |

Table 5. Quantitative evaluation for SauTeg datasets.

| **Models** | **Accuracy** | **Recall** | **Precision** | **F1-Score** |
|---|---|---|---|---|
| MobileNet(v3) [56] | 71.66% | 0.516 | 0.572 | 0.542 |
| ResNet50 [7] | 71.38% | 0.548 | 0.577 | 0.562 |
| ConvNeXt [57] | 74.17% | 0.509 | 0.555 | 0.531 |
| EfficientNet [8] | 75.65% | 0.572 | 0.600 | 0.586 |
| Swin-ViT [12] | 65.34% | 0.373 | 0.459 | 0.411 |
| DeiT [11] | 74.23% | 0.554 | 0.569 | 0.562 |
| CrossViT [49] | 75.34% | 0.505 | 0.532 | 0.519 |
| PVT [13] | 80.72% | 0.628 | 0.663 | 0.645 |
| MPViT [58] | 75.13% | 0.539 | 0.596 | 0.566 |
| MViT [14] | 82.96% | 0.661 | 0.680 | 0.670 |
| **Proposed ROI-ViT (Segmentation)** | **83.49%** | **0.667** | **0.707** | **0.687** |
| **Proposed ROI-ViT (CAM)** | **84.66%** | **0.668** | **0.699** | **0.692** |

## 5.3. ROI generator comparison

In the proposed ROI-ViT system, two types of ROI generators were tested based on segmentation and CAM. In Tables 3–5, the last two rows show that the CAM-based ROI generator obtained better results than the segmentation-based ROI generator. This is because of the image characteristics of the ROI maps. Fig. 7 shows examples of segmented maps and CAMs. In this study, PSPNet[59] was tested for segmentation owing to its good performance. As shown in this figure, the segmented maps appear similar to the B/W images. By contrast, CAMs contain more pixel information. For cross-attention fusion, the input pest images are blended with saliency maps to render the CAMs. In the CAM images, the red colors correspond to salient areas, whereas the blue colors indicate less important areas. The proposed ROI generators successfully determined the initial ROIs. This implies that the proposed ROI generators can provide additional ROI information to the proposed ROI-ViT backbone and teach it to focus on the regions of the feature maps during training. The predicted ROI maps may be incorrect for pest images with complex backgrounds and small sizes. However, the initial ROI maps can be transformed into more accurate maps via the cross-attention module, which is discussed in the next section. Additionally, ROI generation is not the final goal of this study. Therefore, a more detailed evaluation of the ROI generator is unnecessary and is beyond the scope of this study.

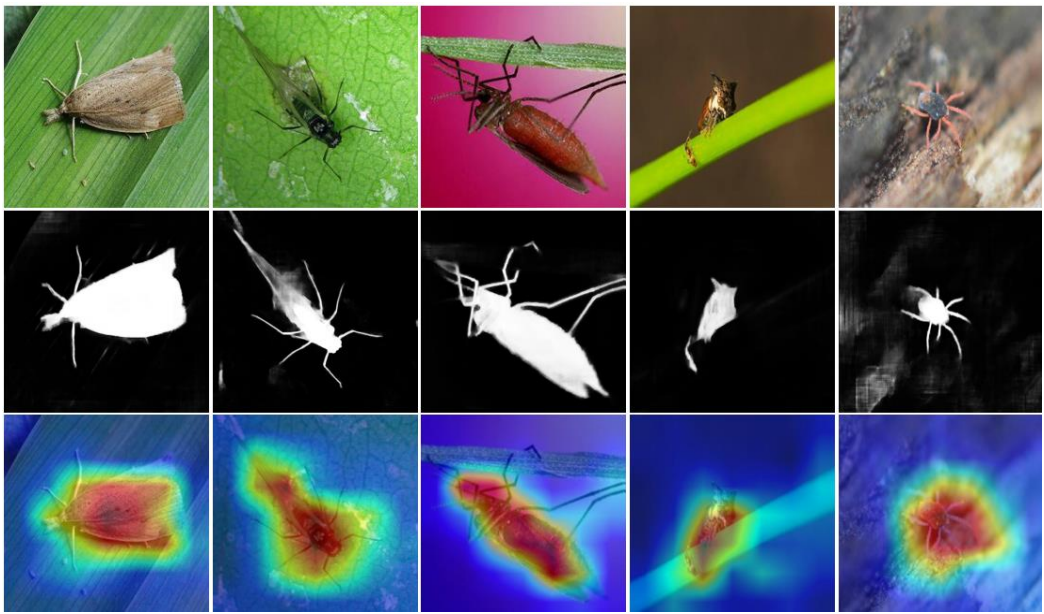

Fig. 7. Experimental results for ROI generators; input pest images (first row), segmented ROI maps (second row), and CAM-based saliency maps (third row).

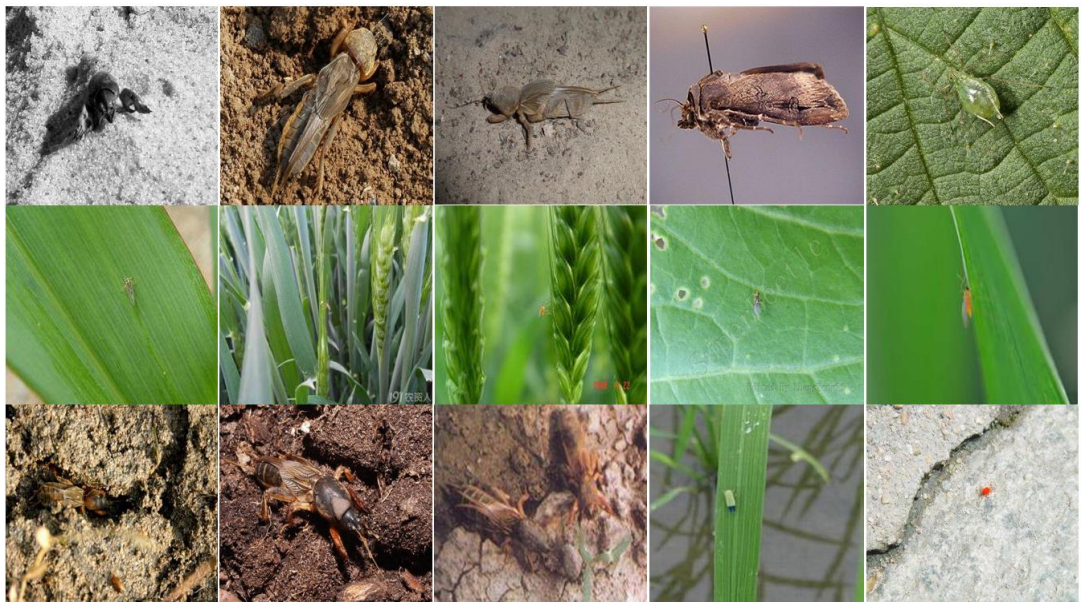

Fig. 8. New test dataset called IP102(CBSS) with complex backgrounds and small sizes.

**Table 6. Quantitative evaluation for the challenging dataset, IP102(CBSS)**

| Models | Accuracy |
|---|---|
| EfficientNet[8] | 64.1% |
| PVT[13] | 60.0% |
| MViT[14] | 58.9% |
| **Proposed ROI-ViT (Segmentation)** | **76.3%** |
| **Proposed ROI-ViT (CAM)** | **76.9%** |

**5.4. Evaluation on a new dataset IP102(CBSS) with complex backgrounds and small sizes**

As emphasized in the introduction with Fig. 1, our motivation starts with the challenging problem of complex backgrounds and small sizes hinder the identification of the initial ROI maps and worsen pest image identification. To prove that the proposed ROI-ViT can overcome this problem, a new test dataset is reconstructed from test dataset of IP102, as shown in Fig. 8. This means that this dataset was used only for testing and not for training. This dataset contained 146 pest images with complex backgrounds and small sizes. We call this dataset IP102(CBSS).

To test the new dataset IP102(CBSS), we selected only four models, namely, MViT, EfficientNet, PVT, and the proposed ROI-ViT, which ranked from 1st to 4th in Tables 3–5. Table 6 lists the accuracies of the IP102(CBSS) dataset. In this Table, it can be seen that there is significant difference in the accuracy between conventional SOTA models and the proposed ROI-ViT. In Tables 3 and 5, the performance of MViT appears to be comparable to that of the proposed ROI-ViT. However, this assumption is incorrect. For the new dataset IP_CBSS, which contained only pest images with complex backgrounds and small sizes, the accuracy of MViT decreased sharply. That is, it decreased by approximately 23%. PVT and EfficientNet also exhibited reduction rates of approximately 16% and 15%, respectively. By contrast, the proposed ROI-ViT can still provide excellent accuracy, although slightly lower, even on the IP102(CBSS). The accuracy decreased by approximately 5%. ***This result confirmed that the proposed ROI-ViT is much more robust to complex backgrounds and small sizes than conventional SOTA models.***

arXiv Version (arxiv.org)arXiv Version (arxiv.org)

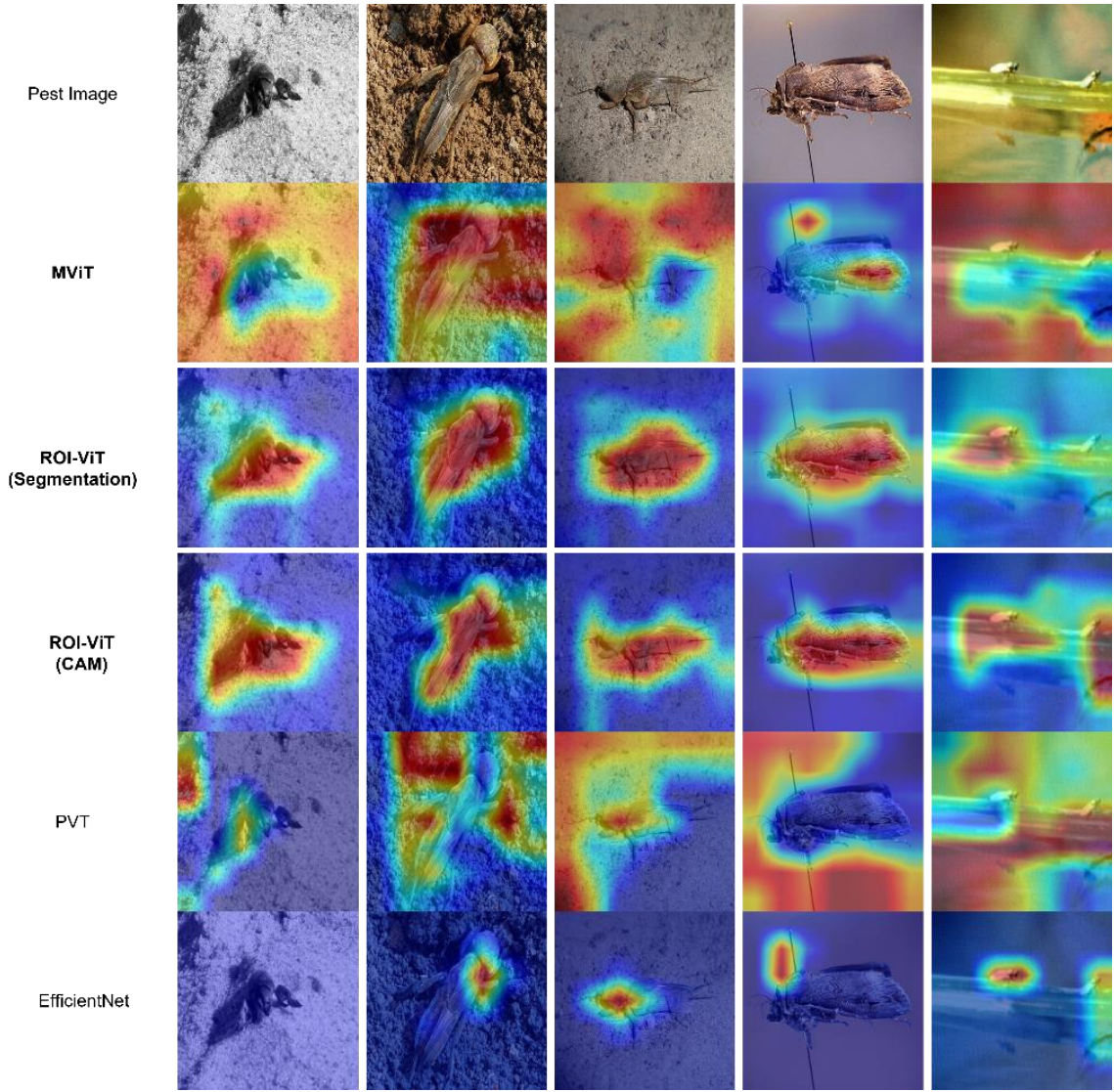

Fig. 9. Visualization of saliency maps for pest images with complex backgrounds and small sizes. Those pest test images are taken from a new challenging dataset IP102(CBSS).

*5.4.1 Proposed cross-attention with ROI correction capability*

We visualized saliency maps using Score-CAM[29] to better understand why the accuracy of MViT decreases sharply, whereas the proposed ROI-ViT maintains its accuracy. Fig. 9 shows visualized saliency maps, which are visual representation learned by MViT, PVT, EfficientNet and proposed ROI-ViT. The first row of Fig. 9 shows examples of input pest images with complex backgrounds and small sizes. The second row shows the saliency maps of MViT, and the third and fourth rows show the saliency maps of the proposed ROI-ViT using segmentation and CAM, respectively. As shown in the saliency maps, MViT failed to learn and detect the ROI through training because of its complex background and small pest sizes. However, the proposed ROI-ViT was successful in finding and updating ROIs even for pest images with complex backgrounds and small sizes. For this reason, the proposed ROI-ViTs can maintain a powerful performance.

In addition, to verify that the proposed multiscale cross-attention fusion can improve the initial ROI maps, the saliency maps of MViT were used as the output of the proposed ROI generator. In other words, the saliency maps in the second row were fed into the proposed ROI-ViT backbone for complementary feature fusion. These saliency maps passed through the proposed ROI-ViT and were transformed into more accurate maps, as shown in the fourth row of Fig. 9. ***This result confirms that our multiscale cross-attention fusion has powerful capability of changing incorrectly estimated saliency map into a better one.***

The saliency maps for Efficient and PVT are provided in the last two rows, as shown in Fig. 9. Note that PVT suffers from separating ROI maps from complex backgrounds, whereas EfficientNet can detect ROI maps more accurately than MViT and PVT. Therefore, EfficientNet has better accuracy than the models for IP102(CBSS), as shown in Table 6.

## 5.5. Ablation study

### 5.5.1 TB and DB numbers

Table 7 presents the accuracy according to the number of TB and DB. In MViT, the number of TB at stage 3 was set to 10. Following this guideline, in the ROI-ViT, the number of TB at stage 3 is first changed while keeping the number of DB fixed. The third row of Table 8 indicates that the highest accuracy was obtained when the number of TB reached 10. The accuracy first increased and then fluctuated as the number of TB increased. Next, the number of TB was fixed at four, second place, and the number of DB was changed from 1 to 10. The last row shows that the accuracy is the highest when the number of DB is eight, but this is lower than when TB = 10 and DB = 1. Therefore, in this study, TB and DB were set to 10 and 1, respectively.

Table 7. Accuracies according to the number of TB and DB

| Stage-3 | TB | 1 | 4 | 6 | 8 | 10 |
|---|---|---|---|---|---|---|
| | DB | 1 | 1 | 1 | 1 | 1 |
| | Accuracy | 79.74% | 80.90% | 80.80% | 80.38% | 81.81% |
| | TB | 4 | 4 | 4 | 4 | 4 |
| | DB | 2 | 4 | 6 | 8 | 10 |
| | Accuracy | 80.94% | 81.18% | 81.27% | 81.63% | 81.56% |

*5.5.2 Model change for Score-CAM generation*

The proposed ROI-ViT may depend on the model $f^c$ used for Score-CAM generation, as shown in Eq. (14). MViT was primarily used as the default because of its excellent performance. However, it is questionable whether a lightweight model can maintain a similar accuracy. Therefore, MobileNet(v3), a lightweight CNN model designed for efficient computation in resource-constrained environments, such as smartphones and other mobile devices, was also tested. The IP102 dataset was used for training and testing. Table 7 shows the accuracy according to the model $f^c$ used for Score-CAM generation. Even though MViT is replaced by MobileNet, the accuracy is almost the same, with a slight drop of 0.19%, and it still outperforms the other SOTA models in Table 3. This indicates that the proposed ROI-ViT method is less dependent on model $f^c$ for Score-CAM generation.

Table 8. Accuracies according to the model $f^c$ for Score-CAM generation

| **Models** | **Accuracy** |
|---|---|
| MViT[14] | 81.81% |
| MobileNet(v3)[56] | 81.62% |

## 6. Conclusions

A challenging problem for pest image identification is that the background is complex, and the objects are small relative to the entire image, which hinders the identification of salient regions and severely reduces recognition accuracy. Even the SOTA models, including MViT, PVT, and EfficientNet, failed to find ROIs, and their accuracies dropped sharply for a new challenging dataset that contained only pest images with complex backgrounds and small sizes. To address this challenging problem, this study presents a novel ViT architecture called ROI-ViT. Specifically, the proposed ROI-ViT was built using a dual branch to leverage the ROI information and progressively update the ROIs and multiscale image features through multiscale hierarchies and cross-attention fusion. To initialize the ROIs, two types of ROI generators were designed based on soft segmentation and CAM, and then incorporated into the proposed ROI-ViT backbone. For multiscale hierarchies, multihead pooling attention was adopted in the transformer block, whereas in the dual block, a new cross-attention fusion was applied to exchange patch tokens from the dual branch and achieve complementary multiscale-feature fusion. Our experiments demonstrated that the proposed ROI-ViT successfully leveraged ROIs and progressively updated them through multiscale hierarchies and cross-attention fusion. For commonly used pest image datasets, the proposed ROI-ViT outperformed the SOTA models such as MViT, PVT, DeiT, Swin-ViT, and EfficientNet. In particular, for the new challenging dataset IP102(CBSS), the proposed ROI-ViT maintained excellent recognition accuracy, whereas the accuracies of the conventional SOTA models decreased sharply. There were significant differences between the accuracies of the proposed ROI-ViT and SOTA models. This confirms that the proposed model is much more robust to complex background and scale problems than the SOTA models.

**Funding**: This work was carried out with the support of Cooperative Research Program for Agriculture Science & Technology Development (Grant no. PJ016303), National Institute of Crop Science (NICS), Rural Development Administration (RDA), Republic of Korea.

**arXiv Version (arxiv.org)**
[50] X. Fu, Q. Ma, F. Yang, C. Zhang, X. Zhao, F. Chang and L. Han, "Crop pest image recognition based on the improved ViT method," Information Processing in Agriculture, Feb. 2023.

[51] Q. Guo, C. Wang, D. Xiao, and Q. Huang, "A novel multi-label pest image classifier using the modified Swin Transformer and soft binary cross entropy loss," Engineering Applications of Artificial Intelligence, vol. 126, Nov. 2023, 107060.

[52] Y. Zhang, L. Chen, and Y. Yuan, "Multimodal fine-grained transformer model for pest recognition," Electronics, vol. 12, no. 12, June 2023, 2620.

[53] X. Wu, C. Zhan, Y.-K. Lai, M.-M. Cheng, and J. Yang, "IP102: A large-scale benchmark dataset for insect pest recognition", In Proc. IEEE Conference on Computer Vision and Pattern Recognition, Long Beach, USA, Jun. 2019, pp. 8787-8796.

[54] https://www.dlearningapp.com/web/DLFautoinsects.htm (accessed 20 Oct. 2023)

[55] https://github.com/SAUTEG/version_1.0 (accessed 20 Oct. 2023)

[56] A. Howard, M. sandler, G. Chu, L. -C. Chen, B. Chen, M. Tan, W. Wang, Y. Zhu, R. Pang, V. Vasudevan, Q. V. Le, and H. Adam, "Searching for MobileNetV3," In Proc. IEEE International Conference on Computer Vision, Seoul, Korea, Oct. 2019, pp. 1314-1324.

[57] Z. Liu, H. Mao, C. -Y. Wu, C. Feichtenhofer, T. Darrell, and S. Xie, "A ConvNet for the 2020s," In Proc. IEEE Conference on Computer Vision and Pattern Recognition, New Orleans, USA, Mar. 2020, pp. 11966-11976.

[58] Y. Lee, J. Kim, J. Willette, and S. J. Hwang, "MPViT: Multi-path vision transformer for dense prediction", In Proc. IEEE Conference on Computer Vision and Pattern Recognition, New Orleans, USA, Dec. 2021, pp. 7277-7286.

[59] H. Zhao, J. Shi, X. Qi, X. Wang, and J. Jia, "Pyramid scene parsing network", In Proc. IEEE Conference on Computer Vision and Pattern Recognition, Honolulu, USA, Jul. 2017, pp. 6230-6239.
35